# Risk & returns around FOMC press conferences: a novel perspective from computer vision [*]

Alexis Marchal[†]

January 7, 2021

I propose a new tool to characterize the resolution of uncertainty around FOMC press conferences. It relies on the construction of a measure capturing the level of discussion complexity between the Fed Chair and reporters during the Q&A sessions. I show that complex discussions are associated with higher equity returns and a drop in realized volatility. The method creates an attention score by quantifying how much the Chair needs to rely on reading internal documents to be able to answer a question. This is accomplished by building a novel dataset of video images of the press conferences and leveraging recent deep learning algorithms from computer vision. This alternative data provides new information on nonverbal communication that cannot be extracted from the widely analyzed FOMC transcripts. This paper can be seen as a proof of concept that certain videos contain valuable information for the study of financial markets.

[*] I am grateful my supervisors Pierre Collin-Dufresne and Julien Hugonnier for their helpful feedbacks. I also thank Oksana Bashchenko and Philippe van der Beck for useful comments.

[†] EPFL and Swiss Finance Institute. Address: EPFL CDM SFI, EXTRA 129 (Extranef UNIL), Quartier UNIL-Dorigny CH-1015 Lausanne, email: alexis.marchal@epfl.ch

# 1. Introduction

Most central banks actively try to shape expectations of market participants through forward guidance. Some of the main objectives being to impact the price of various securities which in turn influences the financing cost of companies or to reduce market volatility during turbulent times. Over the last few years, we have witnessed an explosion of research papers employing machine learning to analyze various documents produced by central banks. The goal is to measure quantitatively how they communicate. This is usually realized by assigning a sentiment score (positive/negative) to the language employed by the bankers using Natural Language Processing (NLP) techniques.

The contribution of this paper is to provide a new method to characterise the complexity of the discussion between reporters and the Chair of the Fed. Instead of analyzing the text documents, I use the video recordings of the FOMC press conferences and introduce a measure of attention exploiting computer vision algorithms. This is based on the simple premise that complex questions from journalists are followed by complex answers from the Chair, which often creates the need to consult internal documents in order to reply. The main idea is to differentiate between two questions asked by reporters, not by studying their text content, but rather by analyzing how does the Chair behave on the video when answering each question. This way, I am able to identify complex discussions by quantifying how often the Chair needs to look at internal documents. This is the key variable that video images are able to provide over other sources of data. I identify the events that involve more complex discussions and show that they have the largest (positive) impact on equity returns and reduce realized volatility. This highlights a mechanism of uncertainty resolution that works as follows. Answers to complex questions resolve more uncertainty than answers to simple questions and this ultimately impacts stock returns, volatility and the equity risk premium around the press conferences.

Macroeconomic announcement days have been substantially discussed in the asset pricing literature which studies how much of the equity risk premium is earned around these events. Savor and Wilson (2013), Lucca and Moench (2015), Cieslak, Morse, and Vissing-Jorgensen (2019), and G. X. Hu et al. (2019) all find that a significant risk premium is earned around macroeconomic announcements. Ernst, Gilbert, and Hrdlicka (2019) argue that if you account for sample selection and day-of-the-month fixed effects, these days are not special and the risk premium is not that concentrated around macroeconomic announcement days. Regardless of the fraction of the equity premium that is earned on those days, there is *some* risk premium that is earned around these events and they reduce *some* uncertainty by disclosing important information to market participants. This alone makes these events an important object of interest for researchers. Together with Beckmeyer, Grunthaler, and Branger (2019) and Kurov, Wolfe, and Gilbert (2020), all of the above mentioned papers revolve around studying the build-up and resolution of uncertainty around macroeconomic announcements. My addition with respect to this literature is to identify *why* some press conferences reduce more uncertainty than others. To this end, I compare stock returns on FOMC press conference days when reporters had a complex discussion with other days when the talks were arguably simpler according to a new measure of attention. This allows me to identify a channel through which the Fed reduces market uncertainty and affects asset prices: by discussing with financial reporters. This implies that the Chair reveals additional information during the Q&A sessions that is not redundant with the pre-written opening statements. My findings are consistent with Kroencke, Schmeling, and Schrimpf (2018) who show that monetary policy affects the pricing of risk by identifying shocks to risky assets that are uncorrelated with changes in the risk-free rate (i.e. "FOMC risk shifts").

This paper also provides a contribution to the literature of machine learning methods used to analyze central banks communication. I quantify the degree of complexity of a discussion without relying on NLP techniques, hence avoiding their limitations.[1] This new alternative dataset of videos allows me to analyze the press events from a new angle. Indeed, I investigate the same events but leverage computer vision to extract a different signal which is the time spent by the Chair reading documents while answering questions. In other words, the NLP literature has focused on *what* is being said during the press conferences while I focus on *how* it is being said. This is accomplished by exploiting the images of the conferences and scrutinizing the human behavior. This information is not present in the transcripts and I argue that it is valuable for financial markets. However, it is likely that the signal I construct using videos could be augmented by sentiment measures extracted from text data. This is why I view my method as complementary to what has been done in the NLP literature. However, the combination of both approaches is left for future research. Another interesting use of machine learning to analyze FOMC conferences is present in Gomez Cram and Grotteria (2020). Their dataset is closely related to mine in the sense that they also use the videos from FOMC press conferences but only analyze the audio in order to match sentences of the Chair with market reactions in real time. In comparison, my paper is the first to use the images from these videos. Overall, I present a proof of concept that FOMC videos actually provide useful information for financial economists.

In accounting, papers like Elliott, Hodge, and Sedor (2012), Blankespoor, Hendricks, and Miller (2017) and Cade, Koonce, and Mendoza (2020) have used video data to analyze the effects of disclosure through videos. However they do not use any systematic algorithm to extract the visual content which makes the approaches hardly scalable. Some authors like Akansu et al. (2017), Gong, Zhang, and Jia (2019) or A. Hu and Ma (2020) use machines to process the videos but they focus on extracting emotions either from CEOs or entrepreneurs. None of their methods are suited to analyze FOMC press conferences because central bankers exerce an effort to appear as neutral as possible when they speak. In contrast to this literature, I develop a simpler tool that systematically computes the reading time of a person appearing in a video. This fully objective measure does not rely at all on emotions.

The rest of the paper is organized as follows. Section 2 establishes the methodology to construct the attention measure. Section 3 presents the main results and finally section 4 concludes. A technical discussion on computer vision algorithms can be found in appendix A.

---

[1] One common drawback of NLP methods in finance/economics is the need to create a dictionary of positive and negative words. The choice of which words belong to which set is somehow subjective. Another problem with more advanced methods is the necessity to label the data which might have to be chosen by the researcher.

## 2. DATASET AND METHODOLOGY

I use the video of each FOMC press conference (available on the Fed website) from their start in April 2011 to September 2020.[2] The market data consists of the time-series of the S&P500 index sampled at a frequency of 1-min. Each press conference can be decomposed into two parts. (i) The first one is an introductory statement in which the Chair reads a pre-written speech, detailing the recent decisions of the Fed. (ii) The second part is a Q&A session between financial reporters and the Fed Chair. I focus solely on the Q&A for the following reasons. Most of the literature analyzing press conferences has focused on the 1$^{st}$ part (with a few rare exceptions) even though the Q&A occupies around 82% of the time of the press conference. Moreover, the unprepared character of the Q&A session means that the behavior of the Chair, when answering questions (whether he needs to read documents to answer questions or not for instance), does bring valuable information that has never been analyzed. Indeed, the Q&A is spontaneous and the Chair did not prepare answers to the reporters' questions. Using this data, the main problems I try to solve are

**H1:** How can we measure the complexity of a question and its associated answer?

**H2:** Do complex discussions contribute more to reduce uncertainty?

In order to answer these questions, I need to characterise the content of the press conferences. As previously explained, the existing literature has done so by assigning a sentiment score to the verbal content by combining text transcripts with some NLP algorithm. The new idea in my paper is to characterise a discussion between a reporter and the Chair of the FOMC, not by analyzing the language but rather by considering how the Chair reacts after being asked a question. To this end, I decide to focus on the following dimension: Does the Chair reply directly or does he read some internal documents in order to provide an answer? This information is available in the videos provided by the Fed but it needs to be extracted and converted into a numerical quantity that can serve as input for statistical inference tools. This is done by employing various computer vision algorithms that are new in finance but have been applied for years to solve engineering problems. In this paper, I focus on the economic mechanisms and the value of the information that can be extracted from this alternative data. Therefore I will keep the discussion of the methodology on a high (non-technical) level and invite the reader to consult appendix A for more details. The need for a technical discussion on computer vision can be (partially) avoided because every image processing can actually be easily illustrated. I will simply visually present the result of every computation by showing an image and what kind of information I extract from it. Given that a video is nothing but a collection of still images, I will use these two words interchangeably.

The first step is to construct facial landmarks $l \in \mathbb{R}^2$ which are certain key points on a human face used to localize some specific regions like the eyes, the mouth, the jaw, etc. They can be visualized in figure A.1 in the appendix. In this paper, they will help me track certain movements of the Fed Chair during the press conferences when he is answering a question. Basically, I want to know every time the Chair is looking at some documents. This is accomplished in two steps. (i) First I extract the identity of the people in every frame. This is done via a technique called deep metric learning that is briefly explained in the appendix. I do not linger on this method because it does not add any economic intuition. This is solely used to filter out images where the Chair appears and disregard the others. (ii) Once I have isolated the frames with the Chair, I will only use the landmarks associated with the eyes. In figure 2.1 we can observe the facial landmarks (black dots) that are specific to them. Each eye is localized by 6 vectors, the so-called landmarks, $(l_1, ..., l_6)$ on the 2D plane that is the picture. When the eyes close and open the landmarks will move, effectively tracking their movements. These points are created using an ensemble of regression trees that is also explained in the appendix. From there I compute a measure of eyes openness. I want to compute a scalar value indicating how open or closed are the eyes at every point in time. For this purpose I use the eye aspect ratio (EAR) developed in Cech and Soukupova (2016). On each still image (i.e. at one instant in time), I compute the EAR for eye $j$ by calculating the $L^2$ norm between the eye landmarks

$$\text{EAR}^j = \frac{\|l_2^j - l_6^j\| + \|l_3^j - l_5^j\|}{2\,\|l_1^j - l_4^j\|} \qquad j \in \{\text{left eye, right eye}\} \qquad (2.1)$$

where $l_1^j, l_2^j, ..., l_6^j$ are the facial landmarks characterizing one eye and depicted on the diagram in figure 2.1. The vertical distances are represented by green arrows and appear at the numerator of the EAR. The horizontal distance (red arrow) serves to normalize. The final EAR is a simple average for both eyes

$$\text{EAR} = \frac{\text{EAR}^{\text{left eye}} + \text{EAR}^{\text{right eye}}}{2}. \qquad (2.2)$$

---

[2] I remove the conference from the 15$^{th}$ of March 2020 simply because there is no video available (it is only audio).

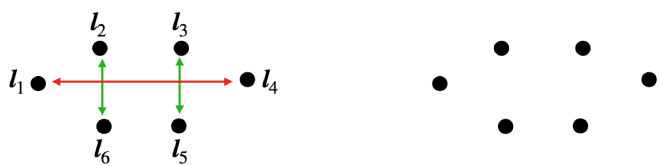

Figure 2.1: Facial landmarks for the left eye and associated distances (colored arrows).

Computing this variable for each frame of the videos means that an EAR scalar value is associated with every single image in my dataset. I denote by $\text{EAR}_{i,t}$ the eye aspect ratio, during the press conference $i$ at instant (frame) $t$. For each FOMC press conference, I obtain a time series of eye aspect ratio for the Chair. An example is provided in the plot of figure 2.2. The blank spaces correspond to times when reporters ask questions and appear on camera. Given that the Chair is not visible during these periods I do not compute an EAR. The scalar quantity EAR provides a simple way to measure if the eyes are open or closed at every point in time. The EAR takes high values when the eyes are wide open and approches 0 as they close. To convince the reader that this variable indeed captures what I intend, I provide an example of two video frames in figure 2.3 in order to compare the EAR in different situations. In figure 2.3a, the Chair Janet Yellen is not looking at the documents on the desk and the associated EAR is 0.33 (relatively high value). The convex hull connecting all the landmarks $l_1, ..., l_6$ (drawn in green around the eyes) creates a relatively large set. In the other picture 2.3b she is clearly reading and the EAR drops to 0.16. Here the convex envelope generates a smaller set. When the Chair spends a substantial amount of time reading, it will produce a series of $EAR_{i,t}$ that will be lower during this time period.

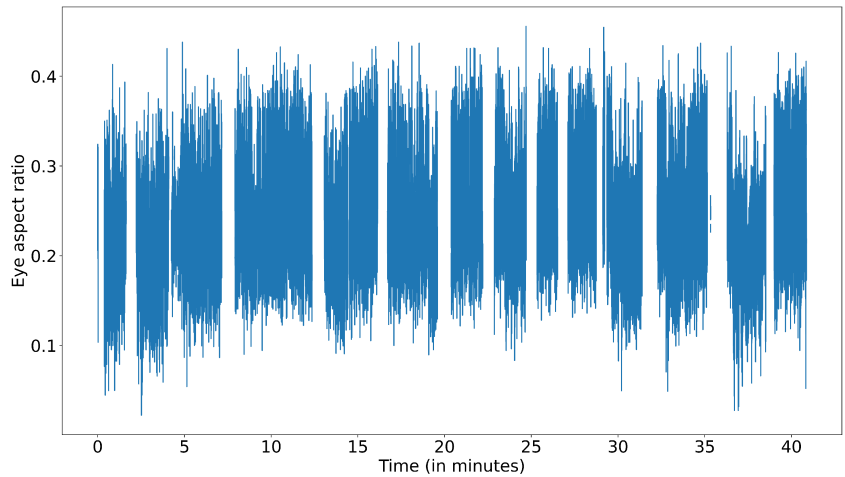

Figure 2.2: Time series of the eye aspect ratio (EAR) of the Fed Chair during the Q&A session of the FOMC press conference (April, 29 2020). The blank spaces correspond to times when reporters ask questions and appear on camera. I therefore do not calculate an EAR during these periods.

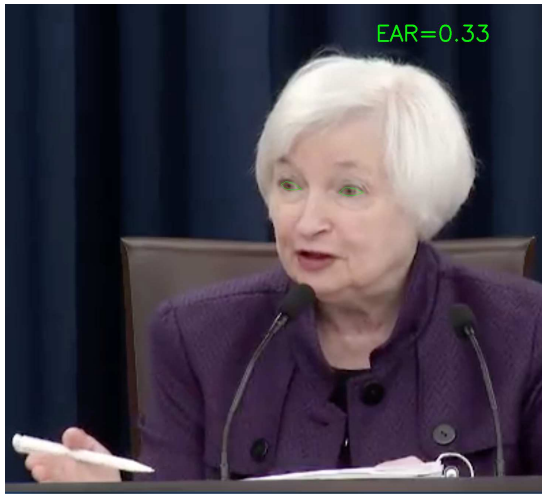 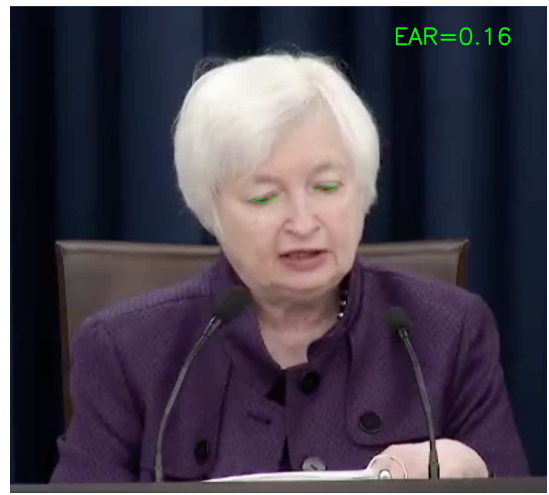

(a) EAR when the Chair is not reading.  (b) EAR when the Chair is reading a document.

Figure 2.3: Convex hulls created by the landmarks $l_1, ..., l_6$ and associated eye aspect ratios (EARs).

It is natural to wonder what type of questions will cause some reading by the Chair. To clarify this, I report below a comparison of two questions asked by reporters. They are copied from the transcript of the press conference of September 21, 2016.

- $Q^*$: **Question from a reporter that does not lead to the consultation of internal documents by the Chair:** "Chair Yellen, at a time when the public is losing faith in many institutions, did the FOMC discuss the importance of today as an opportunity to dispel the thinking that the Fed is politically compromised or beholden to markets?"

- Q': **Question from a reporter that does trigger substential reading from the Chair:** "Madam Chair, critics of the Federal Reserve have said that you look for any excuse not to hike, that the goalposts constantly move. And it looks, indeed, like there are new goalposts now when you say looking for further evidence and-and you suggest that it's evidence that labor-labor market slack is being taken up. Could you explain what for the time being means, in terms of a time frame, and what that further evidence you would look for in order to hike interest rates? And also, this notion that the goalposts seem to move, and that you've indeed introduced a new goalpost with this statement."

The whole idea of this paper is to differentiate between $Q^*$ and Q', not by studying the text content, but by analyzing how does the Chair behave when answering each question. The question Q' will be associated with a complex discussion because my measure of attention EAR will be low due to the reading from the Chair. On the other hand, the EAR stays relatively high when Janet Yellen answers question $Q^*$. For simplicity, I focus solely on where the Chair looks while answering reporters' questions. More sophisticated measures incorporating extra facial landmarks on top of the ones locating the eyes could produce a more precise signal.

So far, for each press conference $i$ I have a time series of EAR. In order to compare the macroeconomic announcements, I decide to summarize the time series information into a variable $\Lambda_i$ that will take one single value per conference. This is done by integrating the EAR over time. I only include the values below some threshold $c$ in order to approximate the total time spent looking at internal documents. The attention measure is therefore defined as

$$\Lambda_i = \int_0^{T_i} \text{EAR}_{i,t} \mathbb{1}_{\{\text{EAR}_{i,t} < c\}} dt \qquad (2.3)$$

where $T_i$ is the time at which the press conference finishes. The constant $c$ helps discriminate when is the EAR low enough for the person to be classified as looking down. The variable $\Lambda_i$ aggregates all the necessary information by measuring how much did the Chair look at his documents in a given press conference. The interpretation of $\Lambda_i$ is as follows. If the value is small, it means that the Chair did not spend much time looking at his documents during conference $i$. If on the other hand the value is large, the Chair spent a significant amount of time looking down. I argue (and show later) that $\Lambda$ is directly proportional to the quantity of uncertainty that has been resolved during a Q&A session. Indeed, the Chair is more likely to look at documents when providing a complex answer. This in turn provides more relevant information to the market and thus reduces uncertainty. It is worth emphasizing that the spontaneity of the questions and answers is important for this analysis. Had the speaker received the questions in advance, this methodology would probably not work.

To conclude the methodology section, it is important to notice that even though I use machine learning methods to extract facial landmarks, the analysis is totally transparent. I would obtain approximately the same variables and results if I had personally watched with great attention all the press conferences and timed manually whenever the Chair is paying attention to the documents in his possession. Machine learning is being used only to automatize this procedure. The variable extracted from the computer vision algorithm $\text{EAR}_{i,t}$ is easily interpretable as an attention measure (i.e. where the person is looking). In the next sections, I will use this data as an input in a linear regression in order to explain the behavior of returns and uncertainty around the FOMC press conferences.

## 3. MAIN RESULTS

In this section I explore how the attention measure of the Chair $\Lambda$ can explain equity returns and quantify the uncertainty that has been resolved due to the conference. It is based on the premise that Q&A sessions involving complex discussions (associated with higher values of $\Lambda_i$) will further reduce uncertainty for financial markets, leading to higher stock returns and lower volatility. Before proceeding further, it is worth noticing that the level of $\Lambda$ is not stationary as illustrated in figure 3.1a. This is the reason I define and work with a new variable

$$\Delta \lambda_i = \lambda_i - \lambda_{i-1} \qquad (3.1)$$

where $\lambda_i = \log(\Lambda_i)$. The new object $\Delta \lambda$ is stationary (see figure 3.1b).

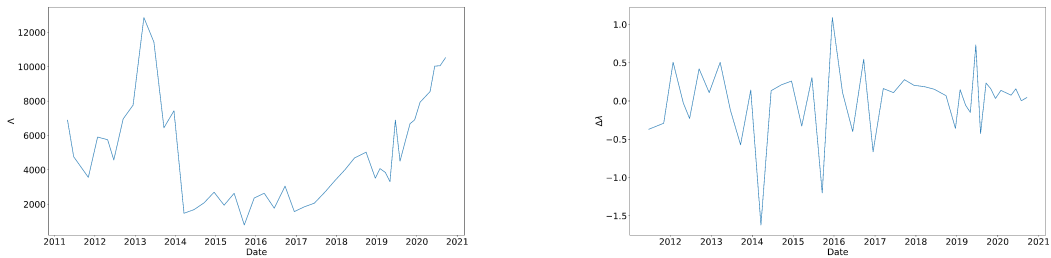

(a) The variable Λ is not stationary.

(b) The log difference of Λ is stationary.

Figure 3.1: Comparison of the level and log difference of the attention measure Λ.

I also construct various benchmark variables that might proxy the complexity of the discussion between reporters and the Fed Chair: # questions, Duration Q&A and Duration speech Chair. The exact construction of these variables is described in appendix B. Each press conference takes place in two stages: (i) an introductory statement and (ii) a Q&A session. Given that I do not analyze the first part, I define only four times $(\tau_i)_{i=1}^4$ that are relevant for the analysis. They are illustrated on the timeline in figure 3.2 and are constructed as follows.

- $\tau_1$ is 2 hours before the beginning of the Q&A session,
- $\tau_2$ is the start of the Q&A session,
- $\tau_3$ is the end of the press conference,
- $\tau_4$ is the end of the trading day.

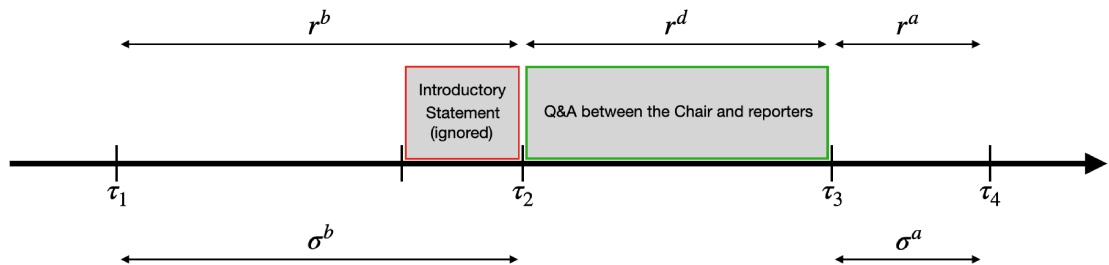

Figure 3.2: Timeline of a typical FOMC press conference.

### 3.1. EXPLAINING CONTEMPORANEOUS STOCK RETURNS

For each press conference $i$, I define the following intraday returns which capture the S&P500 price ($P$) changes respectively during and after the press conference

$$r_i^{\mathrm{d}} = \log\left(\frac{P_{i,\tau_3}}{P_{i,\tau_2}}\right), \tag{3.2}$$

$$r_i^{\mathrm{a}} = \log\left(\frac{P_{i,\tau_4}}{P_{i,\tau_3}}\right). \tag{3.3}$$

I then run a simple linear regression

$$r_i^d = \alpha + \beta \Delta \lambda_i + \varepsilon_i \tag{3.4}$$

of the returns during the Q&A session onto the change in the discussion complexity measure. For comparison I also run similar univariate regressions using the log difference of the benchmark variables. The results are reported in table 3.1. The first observation is that all the betas are positive and of the same order of magnitude. The fact that all variables agree on the sign of the effect is reassuring since they are all measuring the same quantity to some extent. The two most significant variables are $\Delta \lambda$ and the duration of the speech of the Chair (with p-values below 1%). Interestingly, the variable constructed using the video data $\Delta \lambda$ is "better" at explaining equity returns in the sense that its associated t-stat is the highest (around 5) and has by far the highest $R^2$ that is 28.6%. This is not surprising since the duration of the Chair speech is simply measuring how long did the Chair speak during the Q&A session, disregarding anything else. While $\Lambda$ is also proportional to the length of the speech but contains the additional information that the Chair was paying close attention to important documents while answering questions. Hence $\Lambda$ gauges the intricacy of a Q&A session, which is arguably difficult to capture using NLP techniques without making subjective choices (like choosing a dictionary). An interesting consequence of these results is that the Chair reveals additional information during the Q&A sessions that is not present in the pre-written opening statements.

I also run a similar regression to (3.4) but replacing the dependent variable with the return after the conference $r^a$ and find no significant coefficients. This means that the information is incorporated immediately in the stock price over the time of the Q&A.

### 3.2. Impact on uncertainty

The previous section showed that when the change in $\lambda$ is high, stock returns tend to be higher. Given that this variable is independent from any sentiment measure, it is natural to expect that positive returns are caused by a reduction in uncertainty. This is what I argue in this section by showing that $\Delta \lambda$ is negatively correlated with stock market volatility. For this purpose, I simply compute the realized variances before and after each press conference which are denoted by $\sigma_i^b$ and $\sigma_i^a$ respectively. The timeline is illustrated in figure 3.2. The formal construction of

Table 3.1: Explaining contemporaneous stock returns

|  | $r^d$ | | | |
| --- | --- | --- | --- | --- |
|  | (1) | (2) | (3) | (4) |
| const | -0.000 | -0.000 | -0.000 | -0.000 |
|  | (0.001) | (0.001) | (0.001) | (0.001) |
| $\Delta \lambda$ | 0.005*** | | | |
|  | (0.001) | | | |
| $\Delta$# questions | | 0.004 | | |
|  | | (0.002) | | |
| $\Delta$Duration Q&A | | | 0.008** | |
|  | | | (0.004) | |
| $\Delta$Duration speech chair | | | | 0.006*** |
|  | | | | (0.002) |
| Observations | 44 | 44 | 44 | 44 |
| $R^2$ | 0.286 | 0.057 | 0.093 | 0.179 |
| Adjusted $R^2$ | 0.269 | 0.035 | 0.072 | 0.160 |
| Residual Std. Error | 0.004 | 0.004 | 0.004 | 0.004 |
| F Statistic | 16.860*** | 2.554 | 4.322** | 9.161*** |

*Note:* *p<0.1; **p<0.05; ***p<0.01

This table reports the regression statistics for the four different models, each analyzing the explanatory power of a different covariate. The dependent variable $r^d$ is the log return of the S&P500 during the Q&A session (between the beginning until the end). Standard errors are in parenthesis.

these measures is done as follows

$$\sigma_i^b = \sqrt{\frac{1}{\#\mathbb{S}_{\tau_1,\tau_2}} \sum_{\tau=\tau_1}^{\tau_2} r_{i,\tau}^2}, \tag{3.5}$$

$$\sigma_i^a = \sqrt{\frac{1}{\#\mathbb{S}_{\tau_3,\tau_4}} \sum_{\tau=\tau_3}^{\tau_4} r_{i,\tau}^2} \tag{3.6}$$

where $\mathbb{S}_{t,t'}$ is defined as the set containing all the returns between the times $t$ and $t'$ included. Equipped with this measure of change in market uncertainty, I run the main regression of interest

$$\sigma_i^a - \sigma_i^b = \alpha + \beta \Delta \lambda_i + \varepsilon_i. \tag{3.7}$$

Again, for comparison purposes I also replace the regressor with all the log difference of the benchmark variables and report the results in table 3.2. The two most significant variables (with p-values below 5%) are the same than in the previous section: $\Delta \lambda$ and the duration of the Chair's speech. They again both agree on the sign of the effect in the sense that complex discussions reduce market volatility. And consistently with the previous results, $\Delta \lambda$ is the most precise signal in the sense that its associated t-stat is the highest (in absolute) with a value of 2.3. The $R^2$ is also the highest and around 10.9%.

Table 3.2: Explaining the change in volatility before and after the Q&A

|  | $(\sigma^a - \sigma^b) * 100$ | | | |
| --- | --- | --- | --- | --- |
|  | (1) | (2) | (3) | (4) |
| const | -0.002 | -0.002 | -0.002 | -0.002 |
|  | (0.003) | (0.003) | (0.003) | (0.003) |
| $\Delta \lambda$ | -0.014** |  |  |  |
|  | (0.006) |  |  |  |
| $\Delta$# questions |  | 0.002 |  |  |
|  |  | (0.011) |  |  |
| $\Delta$Duration Q&A |  |  | -0.016 |  |
|  |  |  | (0.019) |  |
| $\Delta$Duration speech chair |  |  |  | -0.019** |
|  |  |  |  | (0.009) |
| Observations | 44 | 44 | 44 | 44 |
| $R^2$ | 0.109 | 0.001 | 0.017 | 0.091 |
| Adjusted $R^2$ | 0.087 | -0.023 | -0.006 | 0.069 |
| Residual Std. Error | 0.019 | 0.020 | 0.020 | 0.019 |
| F Statistic | 5.115** | 0.032 | 0.731 | 4.197** |

*Note:* *p<0.1; **p<0.05; ***p<0.01

This table reports the regression statistics for the four different models, each analyzing the explanatory power of a different covariate. The dependent variable $(\sigma^a - \sigma^b)$ is the difference in volatility (realized variation) of the S&P500 before and after the press conference. The time windows over which the realized variation is measured are illustrated in the timeline of figure 3.2. Standard errors are in parenthesis.

## 4. CONCLUSION

This paper develops a new measure of discussion complexity between the Fed Chair and reporters during the Q&A sessions of FOMC press conferences. It is accomplished by analyzing a new dataset of videos and taking advantage of tools from computer vision in order to measure how often the Chair needs to consult internal documents when answering questions. This variable is then showed to explain both contemporaneous equity returns and the change in volatility before and after the conference. On average, complex discussions lead to higher returns and lower volatility. This is consistent with recent findings in the literature that central banks impact the pricing of risk. My work allows to pin down a new mechanism through which press conferences impact the expectations of market participants. A by-product of this result is that there is additional information being revealed during the Q&A sessions that is not redundant with the opening statements. I am currently working on incorporating more data into the analysis by including the videos of the press conferences of the European Central Bank.

In general, my methodology is not constrained to macroeconomic events and can be useful to analyze the nonverbal communication of CEOs or politicians for instance.

# APPENDIX A

This appendix provides a brief overview of the machine learning techniques used in order to compute an eye aspect ratio (EAR) defined in equation (2.1). This paper uses videos as data, however, a video is simply a collection of still images indexed by time and therefore I will always talk about images when referring to the variables used as inputs. Each image $I$ (also called frame) can be represented numerically by a tensor of dimension $\mathcal{D} = x \times y \times 3$. The $x$ and $y$ dimensions respectively correspond to the length and height of the image. Each pixel being an entry in the $x \times y$ matrix that corresponds to its horizontal and vertical location. The 3 represents the three primitive components of any color.

Extracting the behavior of only a subset of people (the Chairs of the FOMC) present in the videos requires using a series of different algorithms one after another. They are all described below in the order used to process the data. My paper employs popular computer vision algorithms for which more details can for instance be found in the book Rosebrock (2017) or the associated online blog[3].

## A.1. IDENTITY DETECTION VIA DEEP METRIC LEARNING

The first task when processing the data is to apply a face recognition algorithm. Meaning that on every single image, I want to know if the Chair of the Fed appears on it or not. For that I need a method that will take as input a frame $I$ and output the identity of the person on it.[4] A powerful tool available is known as deep metric learning.

Let us first begin with a simplified example. Suppose you want to perform a classification of human pictures. That is you want to figure out if there is a human in a given picture or not (without being interested in the identity of the person). This is a simpler task than my original goal but I will build on it later. When one wants to classify labelled images, it is common to train a neural network (typically a convolutional network) that accepts an image as input and outputs a scalar value. For instance the output could be 1 if there is a human face in the picture and 0 otherwise. However, this classification is too simple for me because financial reporters also appear in the FOMC press conference videos and I am not interested in their behavior. This is why instead of using this algorithm, I use a slight modification that will help me filter out the images that do not contain the Fed Chair.

Deep metric learning is different in the sense that the output will be a vector $e \in \mathbb{R}^n$ of embeddings where $n$ is the number of points used to characterize the human face.[5] Formally, our neural network will be a non-linear function $f(.|\theta)$ parametrized by $\theta$ that takes as input a still image $I \in \mathbb{R}^{\mathcal{D}}$ and outputs a vector $e$:

$$f(I|\theta) = e. \tag{A.1}$$

---

[3] https://www.pyimagesearch.com

[4] To simplify the explanations I will assume that in each picture there is only one face. The methods in this appendix do not need this assumption and I do not use it since it does not hold for my dataset.

[5] It is standard to use $n = 128$.

The vector $e$ is describing the face in the picture $I$ in a mathematical way. This step is also called *encoding* the face into a vector. Training the network boils down to making sure the output vectors are close to each other when two pictures of the same person are used as inputs, and far when the persons are different. Suppose that we have two different images $I_1$ and $I_2$ that both contain the same person. We want to train the network such that the associated embedding vectors $e_1$ and $e_2$ are "close". If on the other hand we had two images of different persons, we would like the output vectors to be "far" from each other. In other words, the goal is to twist the network parameters $\theta$ such that two pictures of the same person are classified as having an identical face. For instance any twenty different pictures of the same person should approximatively give the same output vector $e$. In order to perform the training[6], I need to build a new database (different from the FOMC press conferences videos) of $M$ pictures including multiple images of each person I want to detect. Each image $I$ in this set is indexed by $m \in \{1, 2, ..., M\}$. The label is simply the identity of the person in the picture (in my case it is the name of the Fed Chair: Ben Bernanke, Janet Yellen, Jerome Powell). It also helps to include pictures from random people in order to increase the performance of the neural network.

At this stage, one could wonder how does the network embed a face into a real valued vector $e$? For humans it seems natural to compare features like the shape of the eyes, the mouth, the jaws, the length of the nose, etc in order to differentiate people. However, at this step of the process, I do not instruct the machine how to describe a face. I do not specifically constraint it to compare the attributes that seem natural to us humans. The network is learning by itself what are the characteristics that it should pick to properly encode a face. Later, for another task (extracting facial landmarks) I will ask the computer to detect features that are familiar to humans (I will especially be interested in the movement of the eyes).

Once the network is trained and has learnt to properly associate an output vector $e$ to a specific person's face, I generate an embedding vector $e_m$ for every labelled image $I_m$ in my database. Every single face in my database will have an associated mathematical representation that "encodes" it in a vector $e$. It will serve later to compare new unknown faces we want to classify to the labelled images in the database to tell how close they are to each others.

We are now ready to extract the identity of people in the new data. Let $I'$ denote a picture the network has never seen before. By using the following mapping

$$f(I'|\theta) = e' \tag{A.2}$$

we can characterize the face present on $I'$ by using the encoding $e'$. To find out the identity of the person in this new picture I use a simple $k$-NN model (with votes) to make the decision. I measure the distance between $e'$ and all encodings associated with our known faces in our database $e_1, e_2, ..., e_M$. One way to do it is by using the $L^2$ norm

---

[6]In order to save time, I use a pre-trained network that already knows how to encode human faces. I only re-train it on very few images to make sure that it is calibrated for my specific task.

$$d_m = \|\boldsymbol{e}' - \boldsymbol{e}_m\| \tag{A.3}$$

where $d_m$ is the distance between the unknown face in image $\boldsymbol{I}'$ and the known face in image $\boldsymbol{I}_m$. If this distance is small enough (below some tolerance level $\epsilon$), I conclude that both pictures contain the same person. The problem here is that the distance could be small with multiple pictures, not necessarily of the same person (if the encoding is not good for some reason). This is where the voting mechanism enters. To deal with that I create a vector of binary classification $\boldsymbol{a} \in \{0,1\}^M$ whose $m^{\text{th}}$ element in denoted by $a_m$ and is constructed as follows

$$a_m = \begin{cases} 1 & \text{if } d_m < \epsilon, \\ 0 & \text{otherwise.} \end{cases}$$

Basically, $a_m$ contains a yes/no answer to the question: is the person on image $\boldsymbol{I}'$ the same than on image $\boldsymbol{I}_m$? Then I can simply count the number of votes. For instance, for this new image $\boldsymbol{I}'$ I find that it is very close to 40 pictures of B. Bernanke, 2 pictures of Jerome Powell and 1 picture of a random person in my dataset. I will conclude that $\boldsymbol{I}'$ is an image of B. Bernanke. Repeating the above procedure for all the frames in my videos allows me to isolate the times when the Fed Chair appears and disregard moments when the reporters are on the screen.

### A.2. Facial landmarks

After applying the algorithm in the previous section, I now have filtered the parts of the videos that contain the Fed Chair since I was able to extract the identity of a person on an image.

Now I would like to know what is the Fed Chair looking at on every image. However the method used in section A.1 encodes a face in a vector $\boldsymbol{e}$ by selecting facial attributes that do not necessary make sense for us humans.[7] In fact, I do not know what are the exact features selected by the neural network to make the classification. For the previous task it did not matter but now it is a problem because ultimately I want to analyze the position of the eyes of a person. In this section I am going to use an algorithm that can detect face parts that are common to us when describing someone. In technical words I will identify facial landmarks.

A facial landmark is a point represented by a vector $\boldsymbol{l} \in \mathbb{R}^2$ on a 2D image. They are used to localize "important" regions of the face that are present on every human like the eyes, the eyebrows, the nose, the mouth as well as the jaws. An example is displayed in figure A.1. Facial landmarks (green dots) are in fact $(x, y)$-coordinates that map to important regions of the face on $\boldsymbol{I}$. To extract all the landmarks I use a pre-trained facial landmarks detector in the dlib Python library that is based on Kazemi and Sullivan (2014). They rely on an ensemble of regression trees that is trained on a set of human faces. Without entering too much into details, their ensemble method can be thought as a function $g(.|\boldsymbol{\Psi})$ that is parametrized by a vector $\boldsymbol{\Psi}$ and performs the following mapping

---

[7]In order to distinguish two people a human would likely compare the color of the eyes, the length of the nose, the structure of the jaws, etc. However this is not (necessary) how a computer does it.

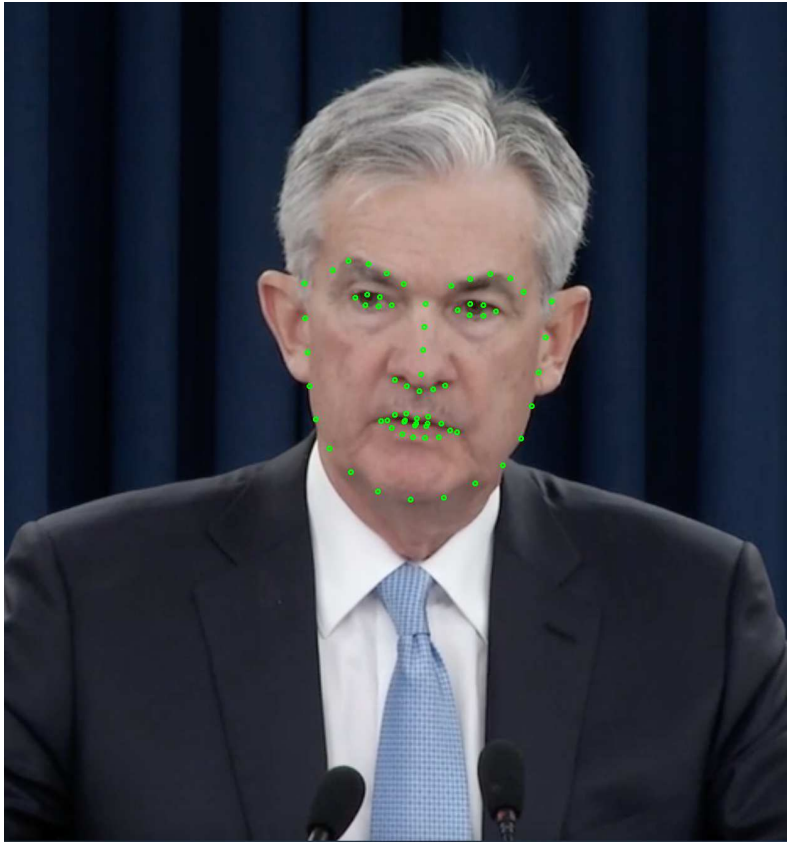

Figure A.1: Picture of Jerome Powell during an FOMC press conference along with all the 68 facial landmarks (green circles) obtained using the pre-trained detector from the dlib Python library.

$$g(I|\Psi) = L. \qquad (A.4)$$

The matrix $L \in \mathbb{R}^{2 \times k}$ collects all the landmark vectors such that

$$L = \begin{pmatrix} l_1 & l_2 & \ldots & l_k \end{pmatrix} = \begin{pmatrix} x_1 & x_2 & \ldots & x_k \\ y_1 & y_2 & \ldots & y_k \end{pmatrix}. \qquad (A.5)$$

Regarding the dimensions of $L$, there are 2 rows because each $l$ is a vector on a 2D image, and $k$ is the number of landmarks we want to extract.[8] Once $L$ is known, I simply keep the landmarks characterizing the eyes and disregard the others. Each eye is is represented by six points (they are drawn in figure 2.1). In order to track the eye movements in my database of videos, I extract a matrix $\boldsymbol{L}$ for every frame $\boldsymbol{I}$ under the constraint that it contains the Chair of the Fed. Then I can easily compute the so called eye aspect ratio (EAR) according to equation (2.1).

---

[8] I use $k = 68$.

In this appendix, I explain the main intuition behind the computer vision algorithms without discussing all the details. The practical implementation of the above algorithms in sections A.1 and A.2 requires some additional steps. One of them is that the user must pre-process the images by first detecting the position of the faces (finding the $(x, y)$-coordinates of a box surrounding the face on image $\boldsymbol{I}$). This can be done in a multitude of ways, including for instance using an other neural network specifically dedicated to that task. In my implementation I use a combination of Histogram of Oriented Gradients (HOG method) and linear Support Vector Machine (SVM). Other small steps are required in order to effectively implement the methods. They can all be found in the above mentioned ressources.

## APPENDIX B

The benchmark variables are constructed as follows

- # questions: The log of the number of questions asked by reporters during a press conference. This is simply constructed by counting the number of interrogation marks in the transcripts.

- Duration Q&A: The log of the time length of the Q&A part of the conference.

- Duration speech Chair: The log of the total time period during which the Fed Chair was speaking. This is constructed using the video data but could also be closely approximated by counting the number of words in the Chair answers in the transcripts.